\documentclass[conference]{IEEEtran}
\IEEEoverridecommandlockouts
\usepackage{amsmath,amsfonts}
\usepackage{epsfig, amssymb}
\usepackage{hyperref}
\usepackage{algorithmic}
\usepackage{algorithm}
\usepackage{array}
\usepackage[caption=false,font=normalsize,labelfont=sf,textfont=sf]{subfig}
\usepackage{textcomp}
\usepackage{stfloats}
\usepackage{url}
\usepackage{verbatim}
\usepackage{graphicx}
\usepackage{cite}
\usepackage{mathtools}
\usepackage{bbold}
\usepackage{ntheorem}
\usepackage{tikz}
\usepackage{pgfplots}
\pgfplotsset{compat=1.18}

\def\R{\mathbb R}

\renewcommand\epsilon{\varepsilon}

\def\vx{{\mathbf{x}}}

\def\vy{{\mathbf{y}}}

\def \vmu{{\boldsymbol{\mu}}}

\def\mI{{\mathbf{I}}}

\def\mX{{\mathbf{X}}}

\def\R{\mathbb{R}}

\newcommand{\ie}{\emph{i.e.,}~}
\pgfmathdeclarefunction{gauss}{2}{%
  \pgfmathparse{1/(#2*sqrt(2*pi))*exp(-((x-#1)^2)/(2*#2^2))}%
}

\def\BibTeX{{\rm B\kern-.05em{\sc i\kern-.025em b}\kern-.08em
    T\kern-.1667em\lower.7ex\hbox{E}\kern-.125emX}}

\newtheorem{theorem}{Theorem}
\newtheorem{assumption}{Assumption}
\newtheorem{remark}{Remark}
\newtheorem{corollary}{Corollary}
\newtheorem{lemma}{Lemma}
    
\begin{document}

\title{Asymptotic Bayes risk of semi-supervised learning with uncertain labeling\\
}

\author{\IEEEauthorblockN{Victor Léger}
\IEEEauthorblockA{\textit{Laboratoire d'Informatique de Grenoble} \\
\textit{Université Grenoble alpes}\\
Grenoble, France \\
victor.leger@univ-grenoble-alpes.fr}
\and
\IEEEauthorblockN{Romain Couillet}
\IEEEauthorblockA{\textit{Laboratoire d'Informatique de Grenoble} \\
\textit{Université Grenoble alpes}\\
Grenoble, France \\
romain.couillet@univ-grenoble-alpes.fr}}

\maketitle

\begin{abstract}
This article considers a semi-supervised classification setting on a Gaussian mixture model, where the data is not labeled strictly as usual, but instead with uncertain labels. Our main aim is to compute the Bayes risk for this model.
We compare the behavior of the Bayes risk and the best known algorithm for this model. This comparison eventually gives new insights over the algorithm.
\end{abstract}


\section{Introduction}

Semi-supervised learning (SSL) is an extension of the conventional supervised learning paradigm by augmenting the (labeled) training data set with unlabeled data, which then “unsupervisably” serve to boost learning performance. SSL has long been considered to be a powerful tool to make use of large amounts of unlabeled data \cite{10.7551/mitpress/9780262033589.001.0001}. 

However, some work also point out the lack of theoretical understanding of these methods \cite{shahshahani1994,Cozman2006RisksOS,BenDavid2008DoesUD}. Even by considering a mere Gaussian mixtures model (which is one of the simplest possible parametric model one could consider for a classification problem), well-known methods such as Laplacian regularization appears to be uneffective to learn from unlabeled data \cite{mai2018random}.

Fortunately, advances in Random Matrix Theory (RMT) has been exploited to design better methods, by proposing fundamental corrections of known algorithms \cite{mai2021consistent}, and even extend them, for instance by considering uncertain labeling \cite{leger2024large}.

Simultaneously, another field of research has focused on analysing Gaussian mixtures model with statistical physics. Such analysis brings an optimal bound for a given problem, meaning that any possible algorithm cannot reach a better performance \cite{lelarge2019asymptotic,nguyen2023asymptotic}. These optimal bounds are a precious tool to understand whether an algorithm has poor performances because of its design or because of the inherent difficulty of the problem it tries to solve.

Therefore, the objectives of this article are twofolds :
\begin{itemize}
    \item Compute the Bayes risk in the case of uncertain data labeling, inspired by the work of \cite{nguyen2023asymptotic}.
    \item Use the knowledge of this optimal bound to further understand the behavior of the algorithm of \cite{leger2024large}, which performances have been proven to be close to the optimal bound.
\end{itemize}

For simplicity reasons, the model presented in this article is a single-task model, but it is worth to note that most of the conclusions remain true in a multi-task setting, as the previous works it is based on are multi-task models \cite{nguyen2023asymptotic,leger2024large}.

The remainder of the article is organized as follows. Section \ref{sec:model} introduces the model, the assumptions and the aim of the next sections. Section \ref{sec:main} states our main theorem, and gives interpretation of this theorem. Section \ref{sec:proof} gives a succinct proof of the main theorem. Finally, Section \ref{sec:sim} displays simulations of both the optimal bound and the algorithm presented in \cite{leger2024large}.

\section{Model and Main Objective}
\label{sec:model}

We consider a semi-supervised binary classification task with training samples $\mX=[\mX_\ell,\mX_u]\in\R^{p\times n}$ which consists of a set of $n_\ell$ labeled data samples $\mX_\ell = \{\vx_{i}\}_{i=1}^{n_\ell}$ and a set of $n_u$ unlabeled data points $\mX_u = \{\vx_{i}\}_{i=n_\ell+1}^n$. Each labeled data point $\mathbf{x}_{i}$ has an associated couple $(d_{i1},d_{i2})$ of \emph{pre-estimated} probabilities that the vector belongs to one class or the other, such that $d_{i1}+d_{i2}=1$. The goal of the classification task is to predict the genuine class of unlabeled data $\mathbf{X}_u$. In this context, we are interested in computing the Bayes risk of the classification task, \ie the minimal classification error achievable for each unlabeled sample $\vx_i$ with the available data :
\begin{equation}
    \inf_{\hat{y}_i} \mathbb{P}(\hat{y}_i \neq y_i)
\end{equation}
where $\hat{y}_i=\mathbb{E}\left[y_i|\mX\right]$ is the label prediction made for the sample $\vx_i$.

\begin{assumption}[On the data distribution]
\label{ass:data_distribution}
The columns of the data matrix $\mX$ are independent Gaussian random variables. Specifically, the data samples $\left(\vx_1,\dots,\vx_n\right)$ are i.i.d.\@ observations such that $\vx_i \in \mathcal{C}_j \Leftrightarrow \vx_i \sim \mathcal{N}(\vmu_j,\mI_p)$ where $\mathcal{C}_j^{t}$ denotes the Class $j$. We assume that the number of data in each class is the same. We further define the quantity $\lambda=\frac{1}{4}\|\vmu_1-\vmu_2\|^2$, which is called the \emph{signal to noise ratio} (SNR).
\end{assumption}

We study our model in a large dimensional setting, where the dimension and the amount of data have the same order of magnitude, which is practically the case with modern data.

\begin{assumption}[Growth Rate]
\label{ass:growth_rate}
As $n\to \infty$ :
\begin{itemize}
    \item $p/n \to c>0$
    \item $n_\ell/n \to \eta$
\end{itemize}
\end{assumption}

With our notations, in a single-task setting, and assuming that the probability couples of labeled data are either $(0,1)$ or $(1,0)$ (\ie the data is labeled with complete certainty), it has been proved in \cite{nguyen2023asymptotic} that under the previous assumptions, as $p \to \infty$, the Bayes risk converges to
\begin{equation}
\label{eq:bayes_mt}
    \mathcal{Q}(\sqrt{q_u}),
\end{equation}
where $\mathcal{Q}(x)=\frac{1}{\sqrt{2\pi}}\int_{x}^{\infty}e^{-\frac{u^2}{2}} \,\mathrm{d}u$, and the couple $(q_u,q_v)$ satisfies the following equations:
\begin{align}
    q_u &= \lambda\frac{\lambda c q_v}{1+\lambda c q_v} \label{eq:overlaps_mt_u} \\
    q_v &= \eta + (1-\eta)F(q_u), \label{eq:overlaps_mt_v}
\end{align}
with $F(q)=\mathbb{E}\left[\text{tanh}(\sqrt{q}Z+q)\right]$, $Z\sim\mathcal{N}(0,1)$.

Our goal in the remainder of this article is to derive an equivalent result in the more general case where data is not labeled with certainty.
Let us define, for each datapoint $\vx_i$, $\epsilon_i=d_{i2}-d_{i1} \in [-1,1]$. This quantity is enough to charaterize the couple of probabilities, as $d_{i1}+d_{i2}=1$. We observe that $|\epsilon|=1$ means that the data is labeled with certainty in a class, while $\epsilon=0$ means that the data is unlabeled.

To get equation \eqref{eq:overlaps_mt_v}, it is needed to compute the quantity $\hat{y}_i = \mathbb{E}\left[y_i|\mX\right]$, which is the estimation of $y_i$ with the available data $\mX$. In the proof (presented in Section \ref{sec:proof}), a trick allows to compute this quantity as a function of $q_u$, and the labeling information is expressed through the prior distribution of $y$.
\begin{itemize}
    \item For data labeled with certainty, it is know that $y_i=-1$ or $+1$, so  the prior is either $\delta(t-1)$ or $\delta(t+1)$.
    \item For unlabeled data, the prior is uniform over $\{-1,+1\}$. Or equivalently, the distribution function is
\begin{equation*}
    \frac{1}{2}\delta(t-1)+\frac{1}{2}\delta(t+1).
\end{equation*}
    \item When the data is not labeled with certainty but with a probability couple $(d_{i1},d_{i2})$, the prior distribution of $y_i$ becomes
\begin{equation}
\label{eq:distrib}
    d_{i1}\delta(t-1)+d_{i2}\delta(t+1)
\end{equation}
\end{itemize}
The following section states our main theorem using this last prior distribution.

\section{Main Results}
\label{sec:main}

\begin{theorem}
\label{th:main}
Under the previous assumptions, as $p \to \infty$,
\begin{itemize}
\item The Bayes risk converges to
\begin{equation}
\label{eq:bayes}
    \mathcal{Q}(\sqrt{q_u}),
\end{equation}
where $\mathcal{Q}(x)=\frac{1}{\sqrt{2\pi}}\int_{x}^{\infty}e^{-\frac{u^2}{2}} \,\mathrm{d}u$.
\item The overlaps $q_u,q_v$ satisfy the following equations
\begin{align}
    q_u &= \lambda\frac{\lambda c q_v}{1+\lambda c q_v} \label{eq:overlaps_u} \\
    q_v &= \lim_{n\to\infty} \frac{1}{n}\sum_{i=1}^n F_{\epsilon_i}(q_u) \label{eq:overlaps_v}
\end{align}
with $F_\epsilon(q)= \mathbb{E}\left[\psi_{\epsilon}(q+\sqrt{q}Z)\right]$, $Z\sim\mathcal{N}(0,1)$ and 
\begin{equation*}
    \psi_\epsilon(t)=\frac{\text{tanh}(t) + {\epsilon}^2\left(1-\text{tanh}(t)-\text{tanh}^2(t)\right)}{1 - {\epsilon}^2\text{tanh}^2(t)}.
\end{equation*}
\end{itemize}
\end{theorem}
A sketch of the proof of Theorem \ref{th:main} is given in Section \ref{sec:proof}.
The function $F_\epsilon$ is similar to the previous function $F$, but the expression of $\psi_\epsilon$ is not easy to understand as it is.

\begin{remark}
    The function $\psi_{\epsilon}$ can be put in the following (more convenient) form :
    \begin{align*}
        \psi_{\epsilon}(t)&=\text{tanh}(t) + \epsilon^2\left(1-\text{tanh}(t)\right) \\
    &-(1-\epsilon^2)(1-\text{tanh}(t))\sum_{k\geq 1} \epsilon^{2k}\text{tanh}^{2k}(t).
    \end{align*}
\end{remark}

\begin{remark}
\label{rq:approx}
    The function $F_\epsilon$ can be approximated by :
\begin{equation}
    \tilde{F}_\epsilon(q)= \mathbb{E}\left[\tilde{\psi}_{\epsilon}(q+\sqrt{q}Z)\right],
\end{equation}
with $Z\sim\mathcal{N}(0,1)$ and
\begin{equation}
    \tilde{\psi}_\epsilon(t)=\text{tanh}(t)+\epsilon^2(1-\text{tanh}(t)).
\end{equation}
\end{remark}

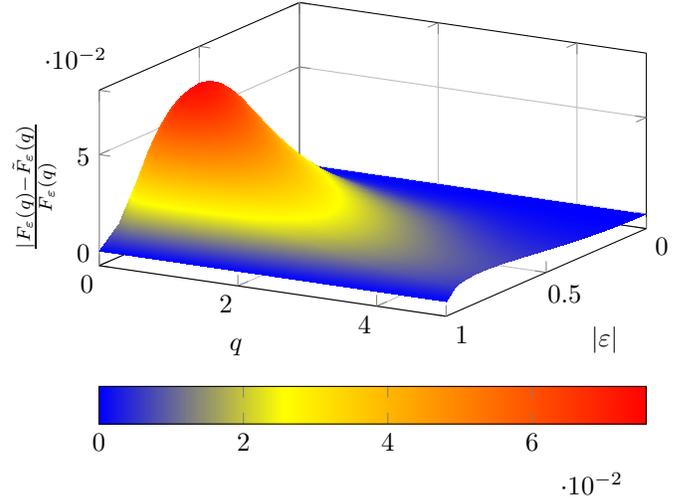
\begin{figure}[!t]
\centering
\begin{tikzpicture}
\begin{axis}[view/h=120, colorbar horizontal, grid=major, yshift=5cm, ylabel={$q$}, xlabel={$|\epsilon|$}, zlabel={$\frac{|F_\epsilon(q)-\tilde{F}_\epsilon(q)}{F_\epsilon(q)}$}, width=1\linewidth, height=0.65\linewidth, legend style={fill opacity=0.8, text opacity=1, font=\small, legend pos=north west}]
    \addplot3 [surf, shader=interp] table {approx_error.dat};
\end{axis}
\end{tikzpicture}
\caption{Relative error of the approximation $\tilde{F}_\epsilon(q)\simeq F_\epsilon(q)$. The error is at most $7\%$, and shrinks for either $\epsilon=0$, $\epsilon=1$ or large $q$}
\label{fig:relative_error}
\end{figure}

Figure \ref{fig:relative_error} gives an idea of the quality of the approximation made in Remark \ref{rq:approx}. However, it is worth to note that its purpose is not to replace the original formula from Theorem \ref{th:main}, as that formula is already tractable and can be computed easily. Instead, Remark \ref{rq:approx} intend to bring a simpler formula that conveys an understanding of the key role of quantity $\epsilon^2$.

As $q_u$ and $q_v$ are related to each other, one could also worry that a small error in equation \eqref{eq:overlaps_v} could lead to a completely different solution for $q_u$ and $q_v$. Fortunately, if one replaces the solution $(q_u^\star,q_v^\star)$ of the system by another solution $(q_u^\star+\Delta q_u,q_v^\star+\Delta q_v)$, then we have $|\frac{\Delta q_u}{q_u^\star}| \leq |\frac{\Delta q_v}{q_v^\star}|$. This means that a small variation of $q_v$ leads to an even smaller variation of $q_u$.

\begin{corollary}
\label{cor:main}
With the previous approximation of the function $F_\epsilon$, one can approximate the equation \eqref{eq:overlaps_v} :
\begin{equation}
\label{eq:qv_final}
    q_v \simeq \bar{\epsilon}^2 + (1-\bar{\epsilon}^2)F(q_u)
\end{equation}
with 
\begin{align*}
    \bar{\epsilon}^2 &= \frac{1}{n}\sum_{i=1} {\epsilon_i}^2 \\
    F(q) &= \mathbb{E}\left[\text{tanh}(q+\sqrt{q}Z)\right] \\
    Z&\sim\mathcal{N}(0,1)
\end{align*}
\end{corollary}

\textit{Proof:} The function $\tilde{F}_{\epsilon_i}$ described in Remark \ref{rq:approx} can be expressed with function $F$ :
\begin{align*}
   \tilde{F}_{\epsilon_i}(q)&=\mathbb{E}\left[\text{tanh}(q+\sqrt{q}Z) + {\epsilon_i}^2(1-\text{tanh}(q+\sqrt{q}Z))\right] \\
    &={\epsilon_i}^2 + \mathbb{E}\left[\text{tanh}(q+\sqrt{q}Z)\right](1-{\epsilon_i}^2)\\
    &={\epsilon_i}^2 + (1-{\epsilon_i}^2)F(q)
\end{align*}
By mixing the results of Theorem \ref{th:main} and Remark \ref{rq:approx}, one gets asymptotically
\begin{align*}
    q_v &\simeq \frac{1}{n}\sum_{i=1}^n \tilde{F}_{\epsilon_i}(q_u) \\
    &=\frac{1}{n}\sum_{i=1}^n \left[{\epsilon_i}^2 + (1-{\epsilon_i}^2)F(q)\right] \\
    &=\bar{\epsilon}^2 + \left(1-\bar{\epsilon}^2\right)F(q)
\end{align*}

Corollary \ref{cor:main} enables an easy interpretation of Theorem \ref{th:main}. Indeed, the value of $q_v$ given by equation \eqref{eq:qv_final} is similar to equation \eqref{eq:overlaps_mt_v}, with $\eta=\bar{\epsilon}^2$. One can check that :
\begin{itemize}
    \item $\epsilon^2=0 \leftrightarrow$ unlabeled data $\leftrightarrow \eta=0$
    \item $\epsilon^2=1 \leftrightarrow$ data labeled with certainty $\leftrightarrow \eta=1$
\end{itemize}
If all samples are labeled with the same value $\epsilon$, then it is equivalent to a task for which one would have a proportion $\epsilon^2$ of data labeled with certainty and a proportion $1-\epsilon^2$ of unlabeled data.

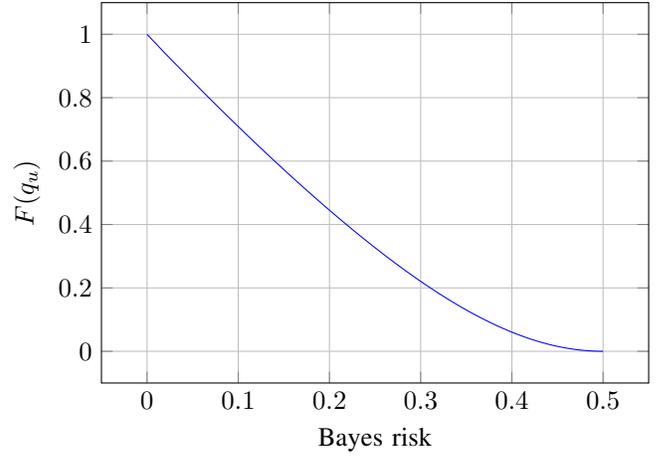
\begin{figure}[!t]
\centering
\begin{tikzpicture}
\begin{axis}[grid=major,yshift=5cm,ylabel={$F(q_u)$},xlabel={Bayes risk},width=1\linewidth,height=0.75\linewidth,legend style={fill opacity=0.8,text opacity=1,font=\small,legend pos=north west}]
    \addplot[blue] table[x=eps,y=y] {usefulness.dat};
\end{axis}
\end{tikzpicture}
\caption{Usefulness of unlabeled data as a function of the Bayes risk of the task. Interestingly, the only criterion to determinate the effectiveness of unlabeled data is how solvable the task is. The lower the Bayes risk is, the more unlabeled data are useful to perform the task.}
\label{fig:usefulness}
\end{figure}

To go further, $F(q_u)$ can be understood as a quantity that expresses how useful unlabeled data are, relatively to labeled data. Indeed, if $F(q_u)=1$, unlabeled data brings as much information as labeled data. Interestingly, this quantity $F(q_u)$ only depends on $q_u$, which itself related to the Bayes risk of the classification task. Thus, usefulness of unlabeled data only depends on how well the task can be performed. Figure \ref{fig:usefulness} displays the quantity $F(q_u)$ as a function of Bayes risk.

\section{Sketch of the proof}
\label{sec:proof}

The proof of Theorem \ref{th:main} is really similar to the one performed in \cite{nguyen2023asymptotic}, as \eqref{eq:bayes_mt} and \eqref{eq:overlaps_mt_u} remain the same, but the main difference lays in the expression of $q_v$, that must be adapted. As in the original proof, $q_v=\langle\hat{\vy},\vy\rangle$ is the \emph{overlap} of the signal $\vy=(y_i)_i$, and we have asymptotically, through the law of large numbers :
\begin{equation}
\label{eq:proof_1}
    q_v = \lim_{n\to\infty} \frac{1}{n}\sum_{i=1}^n \mathbb{E}\left[\hat{y}_iy_i\right]
\end{equation}
where $\hat{y}_i = \mathbb{E}\left[y_i|\mX\right]$ is the MMSE estimator of $y_i$.

Then, a key lemma for the proof is the following.
\begin{lemma}
    Estimating $y_i$ from $\mX$ is asymptotically equivalent to estimating the signal $y_i$ from the output of a Gaussian channel with SNR $q_u$. Let us consider the following Gaussian channel
    \begin{equation*}
        U_i = \sqrt{\lambda}S_i+Z_i
    \end{equation*}
with $\lambda=q_u$ the SNR, $S_i$ the signal and $Z_i\sim\mathcal{N}(0,1)$.
Then computing the overlap $\mathbb{E}\left[\hat{y}_iy_i\right]$ of $y_i$ is equivalent to computing the overlap $\mathbb{E}\left[\hat{S}_iS_i\right]$ of $S_i$, with $\hat{S}_i = \mathbb{E}\left[S_i|U_i\right]$
\end{lemma}

Thus, one has
\begin{equation}
\label{eq:proof_2}
    \mathbb{E}\left[\hat{y}_iy_i\right] = \mathbb{E}\left[\hat{S}_iS_i\right]
\end{equation}

The signal $S_i$ follows the same distribution than $y_i$ :

\begin{equation*}
    S_i \sim \left\{
  \begin{array}{lr}
    -1 ~ \text{with probability} ~ d_{i1},\\
    +1 ~ \text{with probability} ~ d_{i2}
  \end{array}
\right.
\end{equation*}

\begin{equation*}
     \mathbf{E}\left[S_i|U_i\right] = \frac{d_{i1}e^{\sqrt{\lambda}U_i} - d_{i2}e^{-\sqrt{\lambda}U_i}}{d_{i1}e^{\sqrt{\lambda}U_i} + d_{i2}e^{-\sqrt{\lambda}U_i}}
\end{equation*}

Using $\epsilon_i=d_{i1}-d_{i2}$, one gets $\mathbf{E}\left[S_i|U_i\right] = f_{\epsilon_i}(\sqrt{\lambda}U_i)$, with
\begin{equation*}
     f_{\epsilon}(t) = \frac{\text{tanh}(t) + \epsilon}{1 + \epsilon\text{tanh}(t)}
\end{equation*}

\begin{align}
\label{eq:proof_3}
    &\mathbf{E}\left[S_i\mathbf{E}\left[S_i|U_i\right]\right] = \mathbf{E}\left[S_if_{\epsilon_i}(\lambda S_i+\sqrt{\lambda} Z_i)\right] \nonumber \\
    &= d_{i1}\mathbf{E}\left[f_{\epsilon_i}(\lambda+\sqrt{\lambda} Z_i)\right] - d_{i2}\mathbf{E}\left[f_{\epsilon_i}(-\lambda+\sqrt{\lambda} Z_i)\right] \nonumber \\
    &= d_{i1}\mathbf{E}\left[f_{\epsilon_i}(\lambda+\sqrt{\lambda} Z_i)\right] - d_{i2}\mathbf{E}\left[f_{\epsilon_i}(-(\lambda+\sqrt{\lambda} Z_i))\right] \nonumber \\
    &= \mathbf{E}\left[\psi_{\epsilon_i}(\lambda+\sqrt{\lambda} Z_i)\right]
\end{align}
with $\psi_{\epsilon_i}(t) = d_{i1}f_{\epsilon_i}(t) - d_{i2}f_{\epsilon_i}(-t)$

More precisely,
\begin{align*}
    \psi_{\epsilon_i}(t)&=d_{i1}\frac{\text{tanh}(t) + {\epsilon_i}}{1 + {\epsilon_i}\text{tanh}(t)} - d_{i2}\frac{-\text{tanh}(t) + {\epsilon_i}}{1 - {\epsilon_i}\text{tanh}(t)} \\
    &=(d_{i1}+d_{i2})\frac{\text{tanh}(t) - {\epsilon_i}^2\text{tanh}(t)}{1 - {\epsilon_i}^2\text{tanh}^2(t)} \\
    &+(d_{i1}-d_{i2})\frac{{\epsilon_i}(1 - \text{tanh}^2(t))}{1 - {\epsilon_i}^2\text{tanh}^2(t)} \\
    &=\frac{\text{tanh}(t) + {\epsilon_i}^2\left(1-\text{tanh}(t)-\text{tanh}^2(t)\right)}{1 - {\epsilon_i}^2\text{tanh}^2(t)} \\
\end{align*}

Combining \eqref{eq:proof_1}, \eqref{eq:proof_2} and \eqref{eq:proof_3}, we obtain \eqref{eq:overlaps_v}.

\section{Simulations and Applications}
\label{sec:sim}

The objective of this section is to confront theoretical results of Section \ref{sec:main} and the algorithm described in \cite{leger2024large}, which will be from now on refered to as \emph{optimal algorithm}. The common idea of the following experiments is that the optimal algorithm is expected to behave similarly to the optimal bound studied in Section \ref{sec:main}.

As we have seen in Section \ref{sec:main}, different labeling settings can lead to the same value of $\bar{\epsilon}^2$. Let us start with only unlabeled data and data labeled with certainty. Then, 
\begin{equation}
\label{eq:sim1}
    \bar{\epsilon}^2 = \eta,
\end{equation}
and we obtain a classification error $E$.
Now, let us assume that all the labeled data is labeled with the same confidence $\kappa<1$. The total number of data $n$ stays unchanged. For different values of $\kappa$, if one wants to achieve the same error $E$, then more labeled data will be needed as $\kappa$ decreases. Indeed, reaching the same performance means obtaining the same $q_u$, and therefore $q_v$, as the task does not change beyond that. We recall that $q_v \simeq \bar{\epsilon}^2 + (1-\bar{\epsilon}^2)F(q_u)$, and in our context, $F(q_u)$ does not change. Consequently, to obtain the same error $E$, $\bar{\epsilon}^2$ must stay constant. Moreover, we have
\begin{equation}
\label{eq:sim2}
    \bar{\epsilon}^2 = \frac{n_\ell}{n}(2\kappa-1)^2
\end{equation}
By combining \eqref{eq:sim1} and \eqref{eq:sim2}, one gets
\begin{equation}
\label{eq:sim3}
    n_\ell = \frac{\eta}{(2\kappa-1)^2}n
\end{equation}
For a given value of $\eta$, $(2\kappa-1)^2$ must be no smaller than $\eta$, otherwise even a fully labeled dataset would not allow to reach $\bar{\epsilon}^2=\eta$.

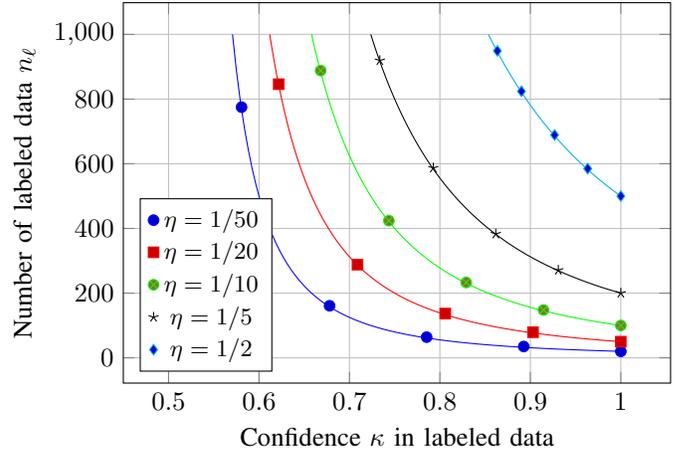
\begin{figure}[!t]
\centering
\begin{tikzpicture}
\begin{axis}[grid=major,yshift=5cm,ylabel={Number of labeled data $n_\ell$},xlabel={Confidence $\kappa$ in labeled data},xmin=0.45,width=1\linewidth,height=0.75\linewidth,legend style={fill opacity=0.8,text opacity=1,font=\small,legend pos=south west}]
    \addplot+[blue, only marks] table[x=conf1,y=nl1] {epsilon_conf_emp.dat};
    \addlegendentry{$\eta=1/50$}
    \addplot+[red, only marks] table[x=conf2,y=nl2]{epsilon_conf_emp.dat};
    \addlegendentry{$\eta=1/20$}
    \addplot+[green, only marks] table[x=conf3,y=nl3] {epsilon_conf_emp.dat};
    \addlegendentry{$\eta=1/10$}
    \addplot+[black, only marks] table[x=conf4,y=nl4] {epsilon_conf_emp.dat};
    \addlegendentry{$\eta=1/5$}
    \addplot+[cyan, only marks] table[x=conf5,y=nl5] {epsilon_conf_emp.dat};
    \addlegendentry{$\eta=1/2$}
    \addplot[blue] table[x=x1,y=y1] {epsilon_conf_th.dat};
    \addplot[red] table[x=x2,y=y2] {epsilon_conf_th.dat};
    \addplot[green] table[x=x3,y=y3] {epsilon_conf_th.dat};
    \addplot[black] table[x=x4,y=y4] {epsilon_conf_th.dat};
    \addplot[cyan] table[x=x5,y=y5] {epsilon_conf_th.dat};
\end{axis}
\end{tikzpicture}
\caption{Number of labeled data $n_\ell$ needed to perform the same performance, as a function of the confidence in the data labeling, for different values of $\eta$ ($n=1000, p=200, \lambda=0.25$). The empirical values are displayed in dots, and theoretical prediction (built on the results of Section \ref{sec:main}) in plain line. The least reliable the data is, the more data is needed to reach the same performance.}
\label{fig:epsilon_conf}
\end{figure}

Figure \ref{fig:epsilon_conf} displays both empirical and theoretical values of $n_\ell$ as a function of $\kappa$ in this setting. The theoretical value is given by \eqref{eq:sim3}, and the empirical one is computed by incrementing $n_\ell$ (and decrementing $n_u$) until the error given by the optimal algorithm gets below $E$. The match between empirical and theoretical curves show that Corollary \ref{cor:main} helps to understand the behavior of the algorithm.

The other takeaway message of Section \ref{sec:main} is that the usefulness of unlabeled data, expressed through the quantity $F(q_u)$, only depends on the Bayes risk of the task, and therefore the final classification error of the optimal algorithm. In order to understand the contribution of unlabeled data, one could be interested in computing the reduction of the classification error by using the semi-supervised version of the optimal algorithm instead of the fully supervised one. With the aim in mind, we will consider the two following quantities:
\begin{itemize}
    \item The absolute error reduction 
    \begin{equation}
        \frac{E_{\text{sup}}-E_{\text{semi-sup}}}{E_{\text{sup}}}
    \end{equation}
    which is the tangible error reduction one can expect by adopting the semi-supervised method instead of the fully supervised one.  
    \item The error reduction relatively to oracle bayes risk
    \begin{equation}
        \frac{E_{\text{sup}}-E_{\text{semi-sup}}}{E_{\text{sup}}-E_{\text{oracle}}}
    \end{equation}
    which reflects how much of the way to oracle error has been done by adopting the semi-supervised method instead of the fully supervised one,
\end{itemize}
where $E_{\text{oracle}}$ is the bayes risk one can expect when the centers of distributions $\vmu_1$ and $\vmu_2$ are known. More precisely, oracle error is given by the formula
\begin{equation}
    E_{\text{oracle}} = \mathcal{Q}(\sqrt{\lambda}).
\end{equation}

\begin{figure}[!t]
\centering
\begin{tikzpicture}
\begin{axis}[grid=major,yshift=5cm,ylabel={Error reduction},ytick={0.0,0.2,0.4,0.6,0.8},yticklabels={$0\%$,$20\%$,$40\%$,$60\%$,$80\%$},xlabel={$\lambda$},width=1\linewidth,height=0.75\linewidth,legend style={fill opacity=0.8,text opacity=1,font=\small,legend pos=south east}]
    \addplot[blue,solid] table[x=lambda,y=algo_abs] {reduction_lambda.dat};
    \addlegendentry{algo/absolute}
    \addplot[red,densely dashed] table[x=lambda,y=algo_oracle] {reduction_lambda.dat};
    \addlegendentry{algo/oracle}
    \addplot[blue,densely dotted] table[x=lambda,y=bound_abs] {reduction_lambda.dat};
    \addlegendentry{bound/absolute}
    \addplot[red,dotted] table[x=lambda,y=bound_oracle] {reduction_lambda.dat};
    \addlegendentry{bound/oracle}
\end{axis}
\end{tikzpicture}
\caption{Percentage of error reduction by using the semi-supervised algorithm instead of the supervised one, as a function of the SNR $\lambda$ ($n=p=200, \eta=0.2$). The easier the task is, the higher the semi-supervised contribution is, because the classification error is lower.}
\label{fig:reduction_lambda}
\end{figure}
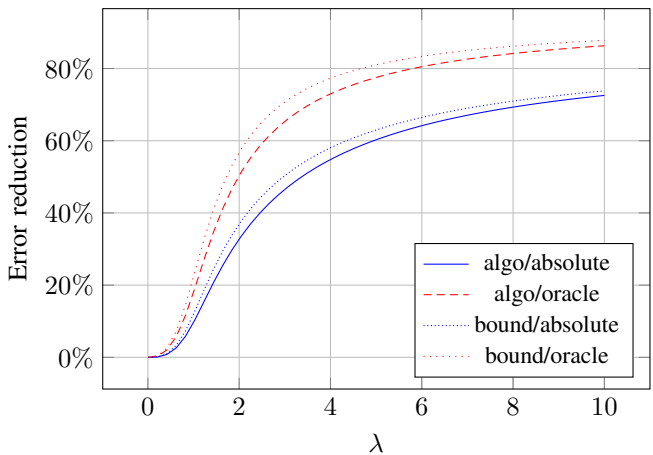

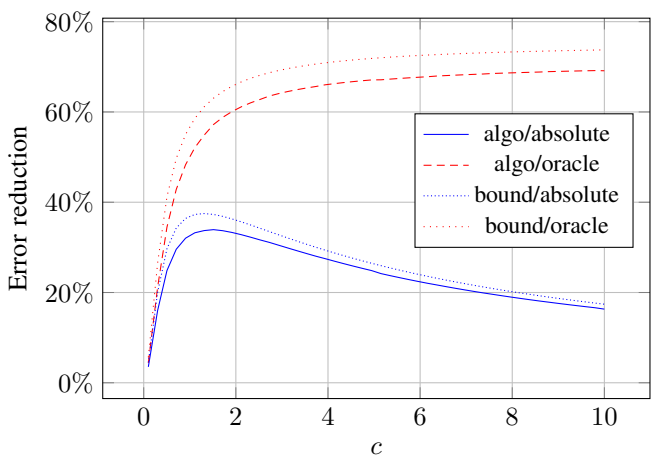
\begin{figure}[!t]
\centering
\begin{tikzpicture}
\begin{axis}[grid=major,yshift=5cm,ylabel={Error reduction},ytick={0.0,0.2,0.4,0.6,0.8},yticklabels={$0\%$,$20\%$,$40\%$,$60\%$,$80\%$},xlabel={$c$},width=1\linewidth,height=0.75\linewidth,legend style={at={(0.97,0.75)},fill opacity=0.8,text opacity=1,font=\small}]
    \addplot[blue,solid] table[x=alpha,y=algo_abs] {reduction_alpha.dat};
    \addlegendentry{algo/absolute}
    \addplot[red,densely dashed] table[x=alpha,y=algo_oracle] {reduction_alpha.dat};
    \addlegendentry{algo/oracle}
    \addplot[blue,densely dotted] table[x=alpha,y=bound_abs] {reduction_alpha.dat};
    \addlegendentry{bound/absolute}
    \addplot[red,dotted] table[x=alpha,y=bound_oracle] {reduction_alpha.dat};
    \addlegendentry{bound/oracle}
\end{axis}
\end{tikzpicture}
\caption{Percentage of error reduction by using the semi-supervised algorithm instead of the supervised one, as a function of the ratio $c=n/p$ ($\lambda=2, p=200, \eta=0.2$). As $c$ grows, the semi-supervised algorithm is more and more effective comparatively to the supervised one, relatively to oracle error, because the classification error is lower and oracle error stays constant. However, if the oracle error ceases to be the reference, then the contribution of semi-supervised decreases for high values of $c$, because both algorithms edge closer to the oracle bound, which stays far from zero.}
\label{fig:reduction_c}
\end{figure}

Leaving out the parameter $\eta$, the main parameters that drive the final error are $\lambda$ and $c$, as we can see in \eqref{eq:overlaps_u}. Therefore, Figures \ref{fig:reduction_lambda} and \ref{fig:reduction_c} display the two kinds of error reduction presented above, respectively as functions of $\lambda$ and $c$.

In Figure \ref{fig:reduction_lambda}, it is clear that the error reduction is higher when $\lambda$ grows, for both types of error reduction. Intuitively, a higher
SNR means a lower final error, and consequently a higher contribution of unlabeled data to the classification.

Meanwhile in Figure \ref{fig:reduction_c}, the two types of error reduction do not behave similarly. In this case, the oracle error is constant, and both errors $E_{\text{sup}}$ and $E_{\text{semi-sup}}$ get close to $E_{\text{oracle}}$ as $c$ grows. Therefore, there is not much to gain by adopting the semi-supervised algorithm, as we are already close to the oracle bound with the supervised one. However, the error reduction relatively to oracle still increases when $c$ grows. We see that the understanding of what remains to be improved plays a key role in Figure \ref{fig:reduction_c}.

\section{Conluding remarks}

Figuring out the link between the performances of an algorithm and its optimal bound gives precious insights. In our case, the algorithm behaves similarly to its optimal bound, giving strong insight that the algorithm is indeed near optimal. So if the algorithm gives poor performances, it simply  means that the problem is inherently too hard to solve. Therefore, the interest of computing such optimal bounds is clear. By knowing in advance how far from optimal an algorithm is, one can avoid spending too much energy to solve a problem which turns out to be a dead-end.

Furthermore, the similarity of behavior between the algorithm and the bound allows to understand the algorithm from an other perspective. Indeed, results from Sections \ref{sec:main} and \ref{sec:sim} provide a new understanding on when semi-supervised learning is truly useful, and when it is not.

\bibliographystyle{IEEEtran}
\bibliography{mybibfile}

\end{document}